\documentclass[10pt,twocolumn,letterpaper]{article}

\usepackage[pagenumbers]{iccv} 

\definecolor{iccvblue}{rgb}{0.21,0.49,0.74}
\definecolor{rise}{RGB}{200,0,0}    
\definecolor{fall}{RGB}{0,0,200}  
\definecolor{SIFT_rise}{RGB}{200,0,0}    
\definecolor{loftr_rise}{RGB}{42, 124, 21}   
\definecolor{loftr_fall}{RGB}{26,92,204}  
\usepackage[pagebackref,breaklinks,colorlinks,allcolors=iccvblue]{hyperref}

\def\paperID{4648} 
\def\confName{ICCV}
\def\confYear{2025}

\title{Monte Carlo Diffusion for Generalizable Learning-Based RANSAC}

\author{
Jiale Wang$^{1*}$ \quad Chen Zhao$^{2}$\thanks{Authors contributed equally}  \quad Wei Ke$^{1}$\thanks{Corresponding author} \quad Tong Zhang$^{23}$ \\
$^1$Xi’an Jiaotong University \quad $^2$EPFL \quad $^3$University of Chinese Academy of Sciences 
}
\usepackage{multirow}

\begin{document}
\maketitle

\begin{abstract}
Random Sample Consensus (RANSAC) is a fundamental approach for robustly estimating parametric models from noisy data. Existing learning-based RANSAC methods utilize deep learning to enhance the robustness of RANSAC against outliers. However, these approaches are trained and tested on the data generated by the same algorithms, leading to limited generalization to out-of-distribution data during inference. Therefore, in this paper, we introduce a novel diffusion-based paradigm that progressively injects noise into ground-truth data, simulating the noisy conditions for training learning-based RANSAC. To enhance data diversity, we incorporate Monte Carlo sampling into the diffusion paradigm, approximating diverse data distributions by introducing different types of randomness at multiple stages. We evaluate our approach in the context of feature matching through comprehensive experiments on the ScanNet and MegaDepth datasets. The experimental results demonstrate that our Monte Carlo diffusion mechanism significantly improves the generalization ability of learning-based RANSAC. We also develop extensive ablation studies that highlight the effectiveness of key components in our framework. The code is released at: \href{https://github.com/comedy0913/MCD}{project page}.
\end{abstract}    
\section{Introduction}
\label{sec:intro}
Robust multi-view geometric estimation is crucial for 3D computer vision tasks such as structure-from-motion~\cite{snavely2008modeling}, SLAM~\cite{mur2015orb}, virtual reality~\cite{szeliski1994image}, and augmented reality~\cite{yu2009real}. The goal is to estimate a reliable geometric transformation model from noisy data with outliers. As the most established robust estimator, RANSAC~\cite{fischler1981random} has been extensively studied and widely adopted for decades. RANSAC operates under the assumption of a parametric model, relying on the consistency with this model to distinguish inliers from noisy data. Therefore, RANSAC is applicable to data generated by any methods for the same task. This property allows RANSAC to integrate seamlessly with multi-view geometry frameworks regardless of the source of the raw data. For instance, in camera pose estimation \cite{hartley2003multiple}, RANSAC is compatible with various pixel-level correspondences, including those obtained from handcrafted detectors/descriptors \cite{lowe2004distinctive,rublee2011orb,bay2008speeded} and learning-based alternatives\cite{detone2018superpoint,sarlin2020superglue,sun2021loftr,edstedt2024roma}. 

\begin{figure}[t]
  \centering
  \includegraphics[width=0.47\textwidth]{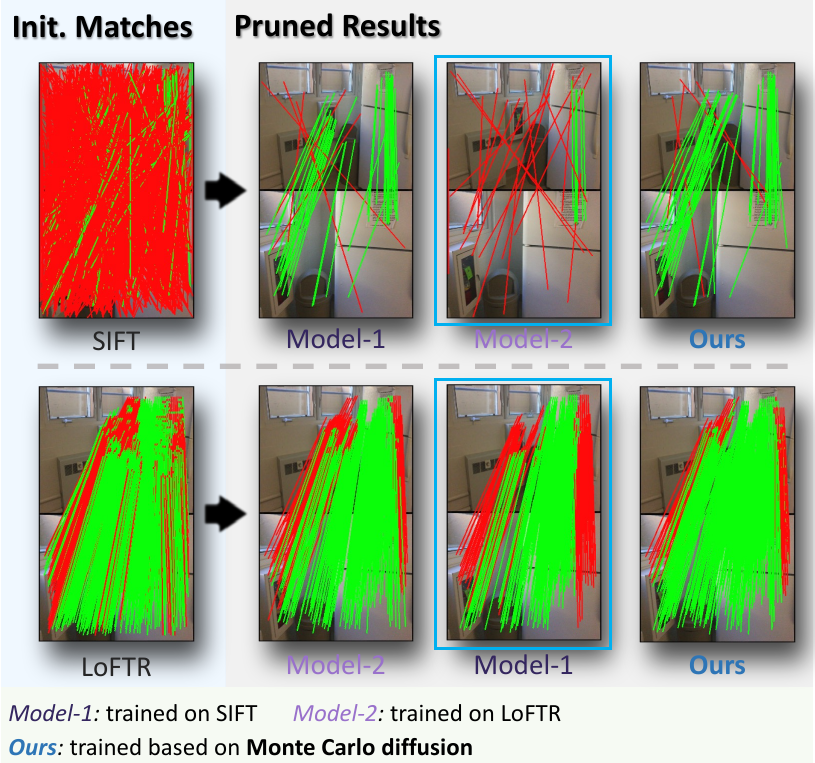} %
   \caption{\textbf{Advantage of Monte Carlo diffusion.} Model-1 and Model-2 denote NG-RANSAC~\cite{brachmann2019neural} trained on SIFT~\cite{lowe2004distinctive} and LoFTR~\cite{sun2021loftr}, respectively. The green lines indicate inliers, and the red ones are outliers. As shown in the blue box, the models trained on specific patterns show limited generalization on out-of-distribution data, e.g., Model-2 trained on LoFTR performs poorly when tested on SIFT. In contrast, we propose a diffusion-based training mechanism where training data is agnostic to specific patterns through a Monte Carlo diffusion process. NG-RANSAC trained on diffused matches demonstrates better generalization across different initial matches. }
   \label{fig:intro}
\end{figure}

The major limitation of RANSAC lies in the sensitivity against outliers. As shown in the literature~\cite{hartley2003multiple,kim2017learned,zhao2020image}, RANSAC becomes less effective when the initial data contains a high proportion of outliers. To handle this issue, some approaches have been proposed to improve the quality of initial data \cite{bian20GMS17gms,yi2018learning,ma2019locality,zhang2019learning, zhao2021progressive}. However, outliers are still inevitable, particularly in challenging scenarios. Therefore, some learning-based RANSAC algorithms \cite{brachmann2017dsac,brachmann2019neural,wei2023generalized} have been introduced, leveraging deep learning to improve the robustness of RANSAC against outliers. In this paper, we focus on investigating these learning-based RANSACs. These methods demonstrate promising effectiveness when tested on in-distribution data. Specifically, in the context of feature matching~\cite{baumberg2000reliable}, the existing learning-based RANSACs are trained and tested on pixel-level correspondences obtained using the \emph{same} algorithm such as SIFT \cite{lowe2004distinctive}. These methods exhibit limited generalization when applied to out-of-distribution data. As shown in~\cref{fig:intro}, the correspondences established via different approaches exhibit significant differences in keypoint positions, spatial patterns, and outlier ratios. A model trained on LoFTR~\cite{sun2021loftr} struggles when applied to SIFT~\cite{lowe2004distinctive} and vice versa, despite both addressing the same geometric problem. This observation indicates that the development of learning-based RANSAC variants \cite{brachmann2017dsac,brachmann2019neural,wei2023generalized} has inadvertently weakened the core strength of RANSAC, limiting its applicability in real-world scenarios where data is typically from diverse algorithms rather than a specific one. 

Consequently, we propose a novel training paradigm, incorporating a diffusion-driven module that eliminates dependence on specific data distributions. Our key insight is to decouple the training process from specific data generation approaches by simulating diverse data patterns. Specifically, we progressively inject the noise to ground-truth data through a diffusion process. To enhance the data diversity, we develop a Monte Carlo sampling mechanism where we introduce different types of randomness at multiple stages of the diffusion module. This stochastic property enables our method to simulate noisy data with diverse distributions. 


We evaluate the applicability of RANSAC in feature matching through comprehensive experiments on ScanNet \cite{dai2017scannet} and MegaDepth \cite{li2018megadepth} datasets that cover diverse indoor and outdoor scenarios, respectively. We utilize SIFT \cite{lowe2004distinctive} and LoFTR \cite{sun2021loftr} to establish pixel-wise correspondences, which represent two types of distributions. Experimental results show that a learning-based RANSAC trained on one distribution fails to generalize to the other during testing, whereas the method trained based on our Monte Carlo diffusion achieves significantly better generalization. Moreover, we conduct ablation studies where the results demonstrate the compatibility of our method with learning-based RANSACs and highlight the effectiveness of key components in our framework. In summary, our primary contributions are threefold: 

\begin{itemize}
    \item We investigate the generalization problem in learning-based RANSAC and identify existing training strategies as the primary factor limiting generalization.
    \item We propose a diffusion-based mechanism that simulates noisy data independent of specific data generation algorithms.
    \item We introduce a Monte Carlo sampling module that enhances the data diversity by injecting multiple sources of randomness at different stages of the diffusion process.
\end{itemize}

\section{Related Work}
\label{sec:formatting}

\begin{figure*}[t]
  \centering
  \includegraphics[width=1.0\textwidth]{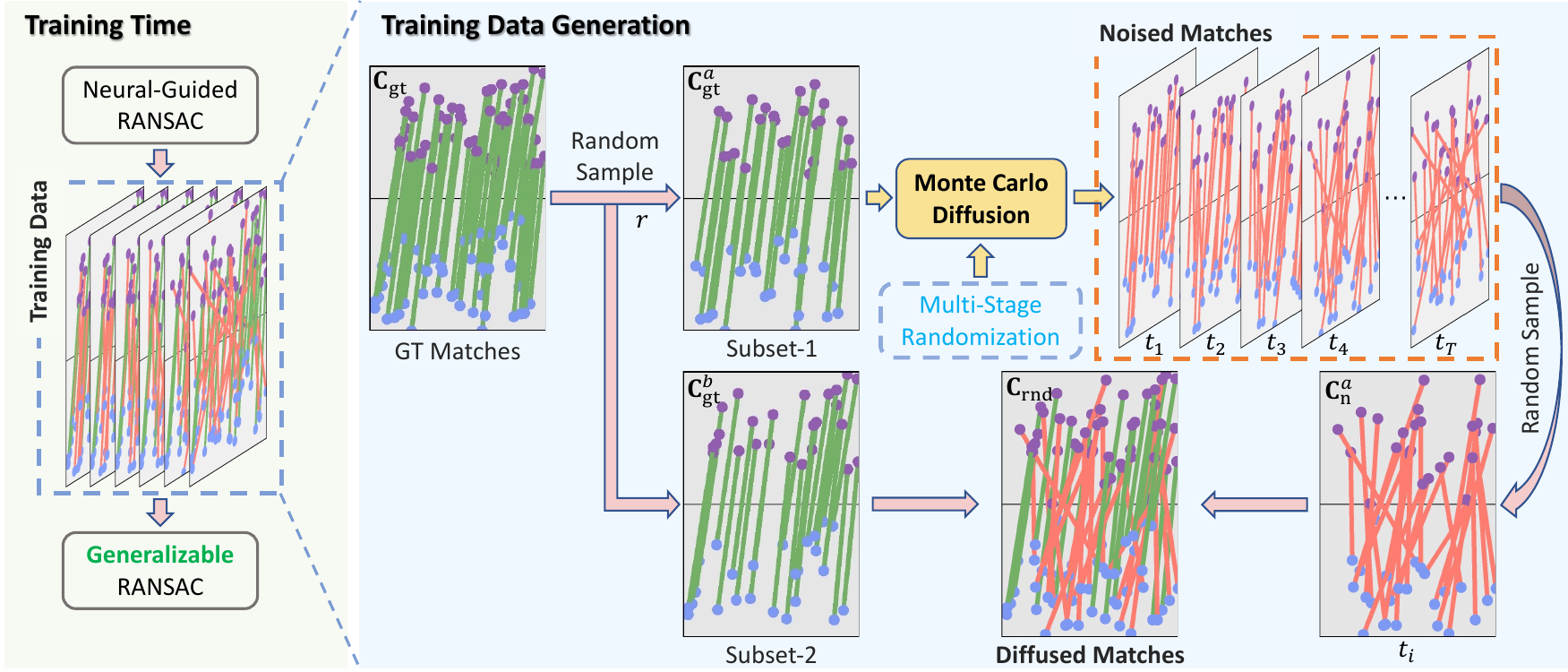} %
   \caption{\textbf{Pipeline of the diffusion process.} We leverage diffusion to simulate noisy data for training learning-based RANSAC. Given ground-truth matches $\mathbf{C}_{\text{gt}}$ between two images, we randomly split them into two subsets $\mathbf{C}_{\text{gt}}^{a}$ and $\mathbf{C}_{\text{gt}}^{b}$. $\mathbf{C}_{\text{gt}}^{a}$ is processed by a Monte Carlo diffusion module with multi-stage randomization, generating multiple sets of noised matches at different timesteps. The final diffused matches are formed by combining $\mathbf{C}_{\text{gt}}^{b}$ as inliers with $\mathbf{C}_n^{b}$ sampled at timestep $t_i$ as outliers. The learning-based RANSAC is then trained on the resulting diffused matches.} 
   \label{fig:pipeline}
\end{figure*}

\noindent\textbf{Random sample consensus.} Random sample consensus (RANSAC) \cite{fischler1981random} has been widely explored in the literature \cite{zhao2020image,Jin2020,hartley2003multiple}, aiming to robustly compute a parametric model from noisy data containing outliers. RANSAC iteratively generates model hypotheses based on subsets randomly sampled from the initial data and assesses the reliability of the generated hypotheses on the initial data. The hypothesis that aligns with the largest number of data points is selected as the most reliable model. The data points consistent with the selected model are identified as inliers. To enhance the effectiveness of RANSAC, several variants have been introduced, such as LO-RANSAC \cite{chum2003locally}, PROSAC \cite{chum2005matching}, USAC \cite{raguram2012usac}, and MAGSAC \cite{barath2019magsac}. These variants modify the key components of RANSAC, such as subset sampling and model verification, based on handcrafted algorithms. The handcrafted RANSAC approaches excel in improving the robustness of model estimation in scenarios where raw data is of sufficient quality. However, the performance deteriorates when a large proportion of outliers exists~\cite{zhao2020image,jin2021image}.
\\

\noindent\textbf{Learning-based robust estimator.}  
To improve the robustness against outliers, recent advances have combined deep neural networks with RANSAC. For instance, some methods, such as~\cite{kim2017learned, zhang2019learning,zhao2019nm,zhao2021progressive}, propose to utilize a network as a pruner to filter out outliers from raw data. The refined data is then processed by RANSAC \cite{fischler1981random} to estimate the parametric model. Notably, these methods act as independent pruners, separate from RANSAC itself. In this context, some approaches develop learning-based alternatives within the pipeline of RANSAC. DSAC \cite{brachmann2017dsac} pioneers this direction by reformulating RANSAC as a differentiable pipeline, enabling end-to-end training of hypothesis scoring networks through gradient-based optimization. Building on this concept, NG-RANSAC \cite{brachmann2019neural} explicitly models a sampling distribution over raw data and employs a network to prioritize high-confidence data points during hypothesis sampling. $\nabla$-RANSAC \cite{wei2023generalized} introduces gradient-driven sampling, which dynamically refines hypothesis proposals using deep feature embeddings. These methods achieve promising performance on in-distribution data during inference. Nevertheless, their training strategy inherently ties them to the statistical properties of specific training data. As we will show in our experiments, existing learning-based RANSACs struggle to generalize to out-of-distribution data without retraining. This limitation weakens the advantage of RANSAC as a general robust estimator in real applications. In this paper, we address this issue by decoupling training from fixed data distributions through a diffusion-driven simulation mechanism.
\\

\noindent\textbf{Pixel-wise feature matching.} Given two images, pixel-wise correspondences are established to compute the geometric transformation for downstream tasks such as image alignment \cite{gao2013seam,brown2007automatic} and 3D reconstruction \cite{yao2018mvsnet,mildenhall2021nerf,kerbl20233d}. Traditional feature matching methods establish correspondences between keypoints based on feature similarities, utilizing handcrafted keypoint detectors and feature descriptors such as SIFT \cite{lowe2004distinctive} and ORB \cite{rublee2011orb}. As shown in the literature \cite{zhao2020image}, these methods often produce numerous false matches, i.e., outliers, particularly in textureless regions and under severe viewpoint changes. In contrast, Learning-based alternatives such as SuperPoint \cite{detone2018superpoint}, SuperGlue \cite{sarlin2020superglue}, and LoFTR \cite{sun2021loftr} employ deep neural networks to generate correspondences, leading to advanced matching quality compared with traditional approaches. Since outliers inevitably exist in initial correspondences for both handcrafted and learning-based methods, RANSAC plays a crucial role in mitigating the impact of outliers and improving the accuracy of model estimation. Moreover, different matching methods produce distinct initial correspondences, characterized by variations in matching density, outlier ratio, and keypoint position. This variability presents significant challenges for learning-based RANSAC \cite{brachmann2019neural,wei2023generalized} when applied to different matching approaches. Unfortunately, the existing learning-based RANSACs exhibit limited generalization to out-of-distribution correspondences. Consequently, we focus on enhancing the generalization of learning-based RANSAC and assessing its impact on feature matching.

\section{Method}
\label{sec:method}
As illustrated in~\cref{fig:pipeline}, our primary innovation lies in employing a diffusion mechanism to progressively transform ground-truth data into noisy variants, with noise intensity increasing over time. Furthermore, we enhance the diversity of the diffused data through Monte Carlo sampling in a multi-stage randomization module. The diffused data points are then used to train a learning-based RANSAC, aiming to robustly estimate the parametric model and identify inliers.

\subsection{Problem Formulation} Note that we focus on the applicability of RANSAC in feature matching. Therefore, we formulate the problem as follows.
Let $\mathcal{M} = \{M_1, M_2, \ldots, M_K\}$ represent the collection of all feature matching methods. For an image pair $(\mathbf{I}, \mathbf{I}^{'})$, each matcher $M_k \in \mathcal{M}$ generates a set of correspondences $\mathbf{C}_k = [\mathbf{c}_1^{(k)}, \dots, \mathbf{c}_N^{(k)}] \in \mathbb{R}^{N \times 4}$, where $\mathbf{c}_i^{(k)} = [x_i, y_i, x'_i, y'_i]$ indicates a correspondence between a keypoint $(x_i,y_j)$ in $\mathbf{I}$ and a keypoint $(x'_i,y'_j)$ in $\mathbf{I}^{'}$. Let $\mathcal{D}_{\text{all}} = \bigcup_{k} \mathcal{D}_{M_k}$ denotes the union of distributions produced by all matchers. The objective of training a learning-based RANSAC is to learn parameters $\theta^*$ that minimize the expected loss over $\mathcal{D}_{\text{all}}$:
\begin{equation}
    \theta^* = \arg\min_{\theta} \mathbb{E}_{\mathbf{C} \sim \mathcal{D}_{\text{all}}} \left[ \mathcal{L}_{\text{RANSAC}}(\mathbf{P}_\theta \mid \mathbf{C}, \mathbf{C}_{\text{gt}}) \right],
\label{eq:final-objective}
\end{equation}
where $\theta$ denotes the learnable parameters in learning-based RANSAC; $\mathbf{P}_\theta$ represents the results predicted from input correspondences $\mathbf{C}$; $\mathcal{L}_{\text{RANSAC}}(\cdot)$ denotes the loss function that will be referred to as $\mathcal{L}(\cdot)$ hereafter for notational simplicity; $\mathbf{C}_{\text{gt}}$ is the ground-truth correspondences.

However, the optimization over $\mathcal{D}_{\text{all}}$ is impractical due to its computational complexity and cost. Instead, existing methods are typically trained on a specific $\mathcal{D}_{M_k}$. They fail to generalize to other $\mathcal{D}_{M_j}\in \mathcal{D}_{\text{all}}, \, j\neq k, $ due to differences in distribution. To overcome this issue, a straightforward solution is to employ multiple matchers for training data generation. Nevertheless, incorporating multiple matchers increases computational cost, and the limited number of matchers lacks sufficient diversity to enable effective generalization. In contrast, we develop a stochastic optimization strategy, approximating $\mathcal{D}_{\text{all}}$ via Monte Carlo sampling. Monte Carlo methods~\cite{rubinstein2016simulation,hammersley2013monte,metropolis1949monte} are grounded in repeated random sampling, excelling in efficiently sampling from complex distributions and sufficiently exploring large solution spaces. In the pipeline of image matching, we integrate Monte Carlo sampling into a match diffusion process, introducing a multi-stage randomization module. At each training iteration, random correspondences $\mathbf{C}_{\text{rnd}}$ are generated based on Monte Carlo sampling. $\mathbf{C}_{\text{rnd}}$, along with the ground truth $\hat{\mathbf{C}}_{gt}$, are then utilized to optimize the learnable parameters in learning-based RANSAC. We reformulate the expected loss over $\mathcal{D}_{\text{all}}$ as an empirical approximation:
\begin{equation}
\label{eq:loss}
\mathbb{E}_{\mathbf{C} \sim \mathcal{D}_{\text{all}}}[\mathcal{L}] \approx \frac{1}{H} \sum_{i=1}^H \mathcal{L}\left(P_\theta \mid \mathbf{C}_{\text{rnd}}^{(i)}, \hat{\mathbf{C}}^{(i)}_{gt}\right),
\end{equation}
where $H$ denotes the number of samplings. 
As the sample size increases ($H \to \infty$), the Monte Carlo estimate of the expected loss converges to the true expectation.

\subsection{Match Diffusion}
To generate $\mathbf{C}_{\text{rnd}}$ in \cref{eq:loss}, we introduce a match diffusion mechanism, injecting random noise into correspondences. Notably, instead of employing diffusion for content generation~\cite{rombach2022high,zhang2023adding,croitoru2023diffusion}, we utilize diffusion to simulate diverse noisy correspondences from available ground truth with varying noise across different timesteps. As the timestep $t\in\{t_1, t_2, \cdots, t_T\}$ increases, the ground truth gradually transitions to pure noise. Specifically, given ground-truth correspondences $\mathbf{C}_{\text{gt}}$ between two images, its noised version at timestep $t$ is generated through the recursive relation:
\begin{equation}
    \mathbf{c}_t^{(i)} = \sqrt{1 - \beta_t} \cdot \mathbf{c}_{t-1}^{(i)} + \sqrt{\beta_t} \cdot \epsilon_{t-1}, \quad \epsilon_{t-1} \sim \mathcal{N}(0, 1),
\end{equation}
where $\mathbf{c}_t^{(i)}$ denotes a correspondence derived from $\mathbf{C}_{\text{gt}}$ at timestep $t$, $\epsilon$ indicates the noise sampled from standard normal distribution, and $\beta_t \in [\beta_{\text{start}}, \beta_{\text{end}}]$ controls the noise injection rate at $t$. We update $\beta_t$ across timesteps based on a linear schedule:  
\begin{equation}
    \beta_t = \beta_{\text{start}} + \frac{t}{T}(\beta_{\text{end}} - \beta_{\text{start}}).
\end{equation}
The recursive formulation is simplified, directly computing $\mathbf{c}_t^{(i)}$ from $\mathbf{c}_0^{(i)}$ as:
\begin{equation}
    \mathbf{c}_t^{(i)} = \sqrt{\bar\alpha_t} \cdot \mathbf{c}_0^{(i)} + \sqrt{1 - \bar\alpha_t} \cdot \epsilon, \quad \epsilon \sim \mathcal{N}(0, 1),
    \label{equ:match-diffusion}
\end{equation}
with $\bar\alpha_t = \prod_{s=1}^t (1-\beta_s)$. As $t$ increases, the injected noise in $\mathbf{c}_t^{(i)}$ grows, causing $\mathbf{c}_t^{(i)}$ to deviate further from the ground truth. Notably, in our method, $\mathbf{c}_0^{(i)}$ represents an inlier, and $\mathbf{c}_t^{(i)}$ serves as an outlier.   

However, diffusing all correspondences in $\mathbf{C}_{\text{gt}}$ may lead to a scenario where the raw data lacks sufficient inliers required for model estimation. Therefore, we introduce a diffusion ratio $r$ to control the proportion of correspondences processed by the diffusion module. As shown in \cref{fig:pipeline}, we randomly sample a subset of ground-truth matches, referred to as $\mathbf{C}_{\text{gt}}^a$, with a ratio of $r$. The remaining matches are denoted as $\mathbf{C}_{\text{gt}}^b$. 
Each correspondence $\mathbf{c}_0^{(i)} \in \mathbf{C}_{\text{gt}}^a$ is perturbed with noise following \cref{equ:match-diffusion}, resulting in the noised counterpart $\mathbf{C}_{n}^a$. In addition, since the noise is sampled from a fixed distribution $\epsilon \sim \mathcal{N}(0, 1)$, similar coordinate shifts may occur in correspondences from $\mathbf{C}_{\text{gt}}^a$ to $\mathbf{C}_{n}^a$. To enhance the diversity, we add a noise scale $s$ during the diffusion process as:
\begin{equation}
    \hat{\epsilon} = \epsilon \cdot s \cdot \max(W, H), \quad \epsilon \sim \mathcal{N}(0, 1),
    \label{equ:scale-factor}
\end{equation}
where $W$ and $H$ represent the image width and height, respectively. \cref{equ:match-diffusion} is then updated as:
\begin{equation}
    \mathbf{c}_{t}^{(i)} = \sqrt{\bar\alpha_t} \cdot \mathbf{c}_0^{(i)} + \sqrt{1 - \bar\alpha_t} \cdot \hat{\epsilon}, \,\,\,\, c_0^{(i)} \in \mathbf{C}_{\text{gt}}^a.
    \label{equ:match-diffusion-updated}
\end{equation}
The final diffused matches are generated by combining noised and clean subsets as:
\begin{equation}
    \mathbf{C}_{\text{rnd}} = \mathbf{C}_{n}^a \cup \mathbf{C}_{\text{gt}}^b, \,\,\,\, \mathbf{C}_{\text{rnd}} \in \mathbb{R}^{N \times 4}.
    \label{equ:conbine-samples}
\end{equation}

\begin{figure*}[t]
  \centering
  \includegraphics[width=0.8\textwidth]{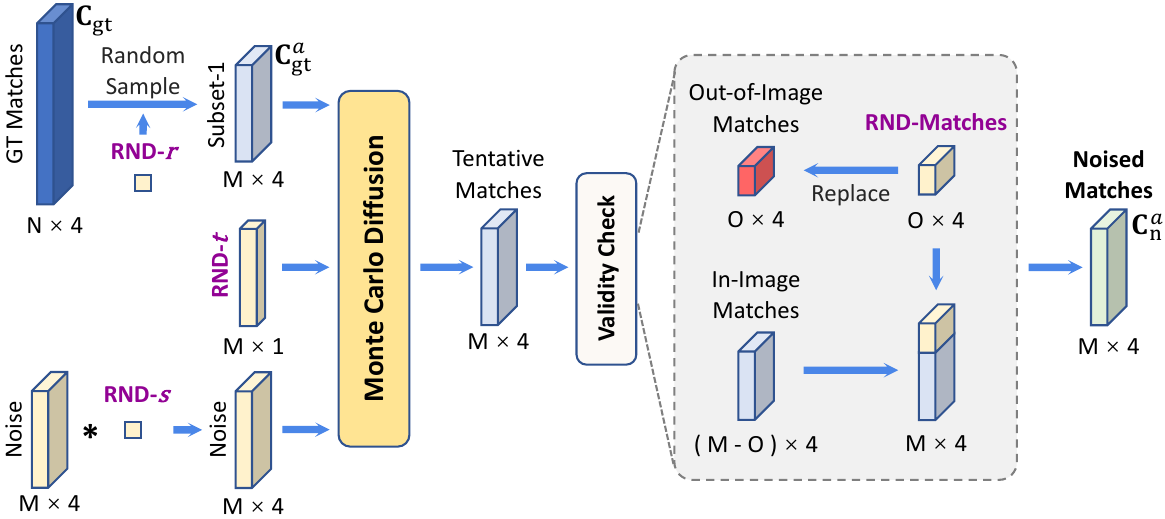} %
   \caption{\textbf{Illustration of the multi-stage randomization module.} We randomly sample the three hyperparameters, timestep $t$, diffusion ratio $r$, and noise scale $s$, in the diffusion mechanism. This multi-stage randomization introduces different sources of randomness into the noised matches, affecting the diffusion intensity, outlier ratio, and noise level, respectively. Invalid matches in the tentative set are replaced by randomly sampled matches, which ensures the validity of the final diffused matches.}
   \label{fig:method}
\end{figure*}

\subsection{Multi-Stage Randomization}
The proposed match diffusion module generates diffused matches controlled by three hyperparameters: timestep $t$, diffusion ratio $r$, and noise scale $s$. Varying these hyperparameters results in different distributions for $\mathbf{C}_{\text{rnd}}$. To determine the hyperparameters, a simple approach is to fix them at predefined values. Nevertheless, such fixed hyperparameters contradict our expectation of Monte Carlo approximation, approximating $\mathcal{D}_{\text{all}}$ based on random sampling. Consequently, to enhance the randomness in the diffusion module, we propose a multi-stage randomization (MSR) method, where we inject randomness at multiple stages throughout the diffusion process.

\begin{figure}[t]
  \centering
  \includegraphics[width=0.5\textwidth]{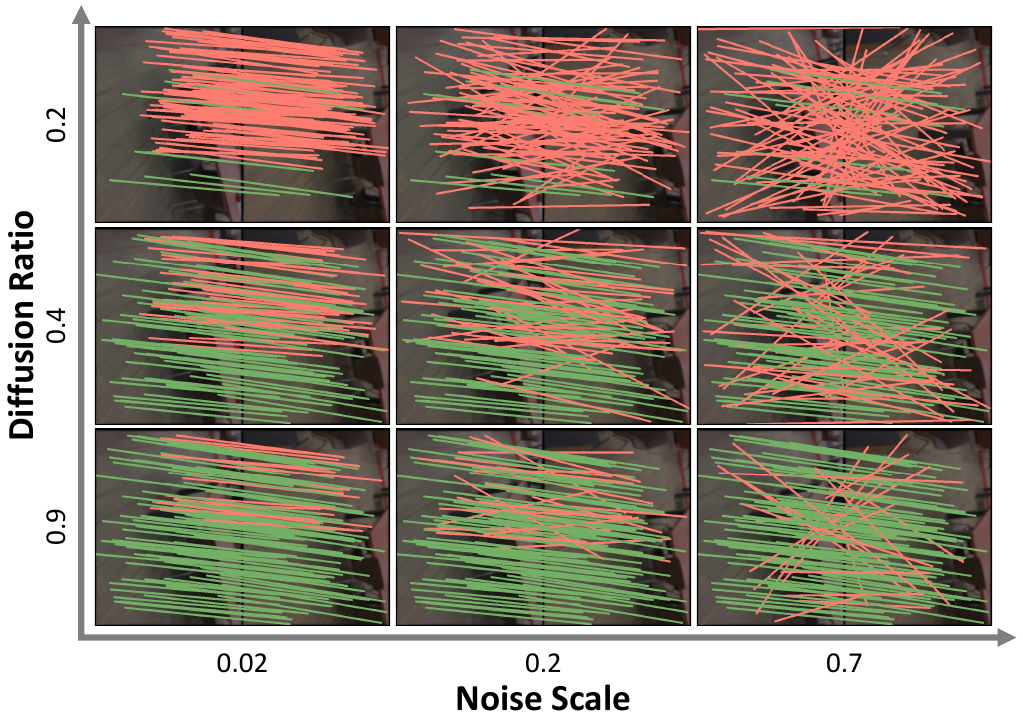} %
   \caption{\textbf{Visualization of diffused matches.} Given the same image pair, different values of diffusion ratio and noise scale result in significantly different diffused matches.}
   \label{fig:Visualization}
\end{figure}

The framework of MSR is illustrated in \cref{fig:method}.
When partitioning $\mathbf{C}_{\text{gt}}$ into $\mathbf{C}_{\text{gt}}^a$ and $\mathbf{C}_{\text{gt}}^b$, we randomly sample ratio RND-$r$ from the range $[r_{\text{min}}, r_{\text{max}}]$ instead of using a fixed $r$. The ratio of noised matches in $\mathbf{C}_{\text{rnd}}$ thereby varies between $r_{\text{min}}$ and $r_{\text{max}}$, leading to diverse outlier ratios.
We then scale the noise $\epsilon$ using a scalar RND-$s$ randomly sampled from the range $[s_{\text{min}}, s_{\text{max}}]$, as formulated in \cref{equ:scale-factor}. Due to the randomness in RND-$s$, the scaled noise $\hat{\epsilon}$ can represent perturbations at varying levels. In addition, for each correspondence in $\mathbf{C}_{\text{gt}}^a$, we randomly sample a timestep, denoted as RND-$t$, from $\{t_1, t_2, \cdots, t_T\}$, and inject the noise to the correspondence as defined in \cref{equ:match-diffusion-updated}.
Notably, during the diffusion process, some correspondences may fall outside the image bounds, making them invalid for training. To handle this problem, we check the validity of the generated correspondences after noise injection and replace invalid matches with $\text{RND-Matches}$. $\text{RND-Matches}$ indicate matches that are regenerated through uniform sampling:
\begin{equation}
 x_{\text{rnd}},x_{\text{rnd}}' \sim \mathcal{U}(0,W), \,\,\,\,
 y_{\text{rnd}},y_{\text{rnd}}' \sim \mathcal{U}(0,H).
 \label{eq:resampling}
\end{equation}
This replacement ensures the validity of noised matches while further increasing the randomness in the diffusion module. By randomly sampling $(\text{RND-r}, \text{RND-s}, \text{RND-t}, \text{RND-Matches})$, our MSR introduces randomness at multiple stages of the diffusion module. \cref{fig:Visualization} illustrates the impact of these hyperparameters on diffused matches. For the same image pair, varying $r$ and $s$ results in significantly different distributions of diffused matches. Therefore, our multi-stage randomization ensures data diversity and thus facilitates effective Monte Carlo approximation. Please refer to the supplementary material for more visualization results.

\section{Experiments}
\subsection{Setup}
We conduct experiments on ScanNet \cite{dai2017scannet} and MegaDepth \cite{li2018megadepth} that include diverse indoor and outdoor scenarios, respectively. We follow the benchmark in SuperGlue \cite{sarlin2020superglue} on ScanNet, splitting the dataset into 1,513 training scenes and 100 test scenes. We randomly sample 20 image pairs per training scene to construct our training set and employ the same image pairs used in SuperGlue \cite{sarlin2020superglue} during inference. 
On MegaDepth, we follow the setup in LoFTR \cite{sun2021loftr}, using 368 scenes for training and 5 scenes for testing. We randomly sample 80 image pairs per scene during training and conduct evaluations on the same image pairs as \cite{sun2021loftr}. 

We employ SIFT \cite{lowe2004distinctive} and LoFTR \cite{lowe2004distinctive} as baselines for feature matching. SIFT represents traditional handcrafted approaches, while LoFTR exemplifies learning-based methods. SIFT is computationally efficient but may produce low-quality matches. In contrast, LoFTR generates dense and accurate correspondences but requires significant computational resources. Both of them have distinct advantages depending on the application and are widely used in 3D computer vision tasks. The evaluation is then performed using NG-RANSAC~\cite{brachmann2019neural} as a learning-based robust estimator by default. Specifically, we train NG-RANSAC separately on SIFT, LoFTR, and our diffused matches. We evaluate the trained models in multiple scenarios, employing AUC of the pose error as a metric~\cite{sun2021loftr,sarlin2020superglue}. Note that our method is also compatible with other learning-based RANSACs, as will be evidenced in our experiments.

\subsection{Implementation Details}
\noindent\textbf{Dataset construction.} For each image pair $(\mathbf{I}, \mathbf{I}^{'})$, the corresponding depth maps and camera parameters are utilized to reconstruct 3D points~\cite{hartley2003multiple}. These points are aligned in the world coordinate system, resulting in the ground-truth pixel-wise correspondences $\mathbf{C}_{\text{gt}}$. We randomly subsample them to obtain 2000 correct correspondences per image pair.

\noindent\textbf{Training details.} We set the diffusion parameters to $\beta_{\text{start}} = 0.0005$, $\beta_{\text{end}} = 0.0025$, and $T = 500$. RND-$r$ and RND-$s$ are randomly sampled within the ranges $[0.2, 0.9]$ and $[0.02, 0.7]$, respectively, resulting in diffused matches with diverse inlier ratios and noise levels. We maintain the hyperparameters in learning-based RANSAC at their default settings to ensure fair comparisons in our benchmarks. We train the model on a GTX 2080 Ti, and the training process takes 60 hours. 

\subsection{Generalization to Out-of-Distribution Data}
\label{sec:ood}
\begin{figure}[t]
  \centering
  \includegraphics[width=0.47\textwidth]{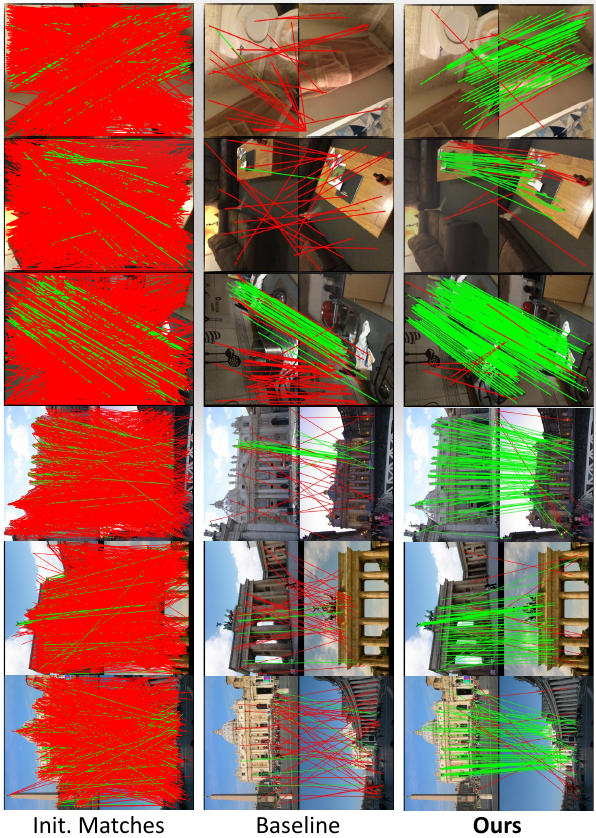} %
   \caption{\textbf{Qualitative results.} Init. Matches represent the initial correspondences generated by SIFT. Baseline and Ours indicate the pruned results using NG-RANSAC trained on LoFTR and diffused matches, respectively. The green and red lines denote inliers and outliers. The baseline shows limited generalization to SIFT, which serves as out-of-distribution data, leading to many outliers after the pruning. In contrast, our method achieves significantly better generalization, identifying more inliers.}
   \label{fig:Qualitative}
\end{figure}

To assess the generalization to out-of-distribution (OOD) data, we conduct experiments on SIFT \cite{lowe2004distinctive} and LoFTR \cite{sun2021loftr} correspondences. Specifically, NG-RANSAC \cite{brachmann2019neural} trained on SIFT is evaluated on LoFTR, and vice versa. Notably, the presented Monte Carlo diffusion (MCD) is agnostic to specific matchers. Therefore, for NG-RANSAC trained on diffused matches, we directly test it on SIFT and LoFTR. \Cref{tab:ng_ransac_scannet} and \Cref{tab:ng_ransac_MegaDepth} list the results on ScanNet \cite{brachmann2019neural} and MegaDepth \cite{li2018megadepth}, respectively. The models trained on the diffused matches obtained through Mante Carlo match diffusion exhibit significant improvements in generalization ability. For instance, on ScanNet, MCD improves AUC $@20^\circ$ by $12\%$ from $48.8\%$ to $60.8\%$ on LoFTR, when compared with the model trained on SIFT. Notably, NG-RANSAC trained on LoFTR yields limited AUCs on SIFT, e.g., $8.5\%$ in AUC $@20^\circ$. Our method significantly enhances the generalization in this case, improving AUC $@20^\circ$ by $17.7\%$ from $8.5\%$ to $26.2\%$. Additionally, we achieve consistent improvements on MegaDepth, increasing AUC $@20^\circ$ by $5.7\%$ and $23.4\%$ when tested on LoFTR and SIFT, respectively. Some qualitative results are illustrated in \cref{fig:Qualitative}.

\begin{table}[htbp]
  \centering
  \resizebox{1.0\linewidth}{!}{
      \begin{tabular}{@{}lllll@{}}
        \toprule
        Training & Testing & AUC $@5^\circ$ & AUC $@10^\circ$ & AUC $@20^\circ$ \\
        \midrule
        LoFTR \cite{sun2021loftr} & SIFT \cite{lowe2004distinctive} & 1.5 & 4.4 & 8.5 \\
        MCD & SIFT \cite{lowe2004distinctive} & \textbf{7.6} {\color{rise}\footnotesize(+6.1)} & \textbf{16.2} {\color{rise}\footnotesize(+11.8)}& \textbf{26.2} {\color{rise}\footnotesize(+17.7)}\\
        \hline
        SIFT \cite{lowe2004distinctive} & LoFTR \cite{sun2021loftr} & 13.0 & 29.7 & 48.8 \\ 
        MCD & LoFTR \cite{sun2021loftr} & \textbf{22.4} {\color{rise}\footnotesize(+9.4)} & \textbf{42.7} {\color{rise}\footnotesize(+13.0)}& \textbf{60.8} {\color{rise}\footnotesize(+12.0)}\\ 
        \bottomrule
      \end{tabular}
  }
  \caption{\textbf{Generalization to out-of-distribution data on ScanNet \cite{dai2017scannet}.} MCD indicates the diffused matches generated via our Monte Carlo diffusion. NG-RANSAC is separately trained on SIFT, LoFTR, and MCD. NG-RANSAC trained on SIFT is evaluated on LoFTR, and vice versa. AUCs of the pose error with different thresholds are reported, and the best results are highlighted in bold.}
  \label{tab:ng_ransac_scannet}
\end{table}

\begin{table}[htbp]
  \centering
  \resizebox{1.0\linewidth}{!}{
      \begin{tabular}{@{}lllll@{}}
        \toprule
        Training & Testing & AUC $@5^\circ$ & AUC $@10^\circ$ & AUC $@20^\circ$ \\
        \midrule
        LoFTR \cite{sun2021loftr} & SIFT \cite{lowe2004distinctive} & 3.1 & 6.7 & 13.4 \\
        MCD & SIFT \cite{lowe2004distinctive} & \textbf{16.9} {\color{rise}\footnotesize(+13.8)} & \textbf{26.2} {\color{rise}\footnotesize(+19.5)} & \textbf{36.8} {\color{rise}\footnotesize(+23.4)}\\ 
        \hline
        SIFT \cite{lowe2004distinctive} & LoFTR \cite{sun2021loftr} & 41.8 & 59.0 & 73.5 \\ 
        MCD & LoFTR \cite{sun2021loftr} & \textbf{53.2} {\color{rise}\footnotesize(+11.4)} & \textbf{67.7} {\color{rise}\footnotesize(+8.7)}& \textbf{79.2} {\color{rise}\footnotesize(+5.7)} \\ 
        \bottomrule
      \end{tabular}
  }
 \caption{\textbf{Generalization to out-of-distribution data on MegaDepth \cite{li2018megadepth}.} NG-RANSAC trained on SIFT, LoFTR, and our MCD is evaluated on out-of-distribution data measured by AUCs of the pose error.}
  \label{tab:ng_ransac_MegaDepth}
\end{table}

\subsection{Comparisons in In-Distribution Scenarios}
\label{sec:id}

In some scenarios, conducting inference on in-distribution data is practical. For example, LoFTR can be deployed in a system with enough computational resources for both training and testing. To assess the applicability of our method in such a setting, we compare our method with NG-RANSAC that is trained and tested on the same matcher. More specifically, we train NG-RANSAC on SIFT and test it on SIFT, and then repeat this process on LoFTR. In contrast, for our method, we retain the diffused matches during training and test the trained NG-RANSAC on both SIFT and LoFTR. As reported in \Cref{tab:ng_ransac_scannet_performance} and \Cref{tab:ng_ransac_megadepth_performance}, our method achieves performance comparable to the baseline, where NG-RANSAC is trained and tested on the same matchers. This evidences that our method not only shows superior generalization but also remains highly competitive in scenarios where the same matcher can be applied for both training and testing.

\begin{table}[htbp]
  \centering
  \resizebox{1.0\linewidth}{!}{
      \begin{tabular}{@{}lllll@{}}
        \toprule
        Training & Testing & AUC $@5^\circ$ & AUC $@10^\circ$ & AUC $@20^\circ$ \\
        \midrule
        SIFT \cite{lowe2004distinctive} & SIFT \cite{lowe2004distinctive} & 7.6 & 16.1 & 25.5 \\
        MCD & SIFT \cite{lowe2004distinctive} & \textbf{7.6} & \textbf{16.2} & \textbf{26.2}\\
        \hline
        LoFTR \cite{sun2021loftr} & LoFTR \cite{sun2021loftr} & \textbf{22.7} & 42.7 & 60.1 \\ 
        MCD & LoFTR \cite{sun2021loftr} & 22.4 & \textbf{42.7} & \textbf{60.8} \\ 
        \bottomrule
      \end{tabular}
  }
  \caption{\textbf{Comparisons in in-distribution scenarios on ScanNet \cite{dai2017scannet}.} The baseline trains and tests NG-RANSAC on the same matchers, while our method trains NG-RANSAC only using diffused matches.}
  \label{tab:ng_ransac_scannet_performance}
\end{table}

\begin{table}[htbp]
  \centering
  \resizebox{1.0\linewidth}{!}{
      \begin{tabular}{@{}lllll@{}}
        \toprule
        Training & Testing & AUC $@5^\circ$ & AUC $@10^\circ$ & AUC $@20^\circ$ \\
        \midrule
        SIFT \cite{lowe2004distinctive} & SIFT \cite{lowe2004distinctive} & 16.6 & \textbf{27.0} & \textbf{38.8} \\
        MCD & SIFT \cite{lowe2004distinctive} & \textbf{16.9} & 26.2 & 36.8 \\
        \hline
        LoFTR \cite{sun2021loftr} & LoFTR \cite{sun2021loftr} & \textbf{53.3} & 67.5 & \textbf{79.3} \\ 
        MCD & LoFTR \cite{sun2021loftr} & 53.2 & \textbf{67.7} & 79.2 \\ 
        \bottomrule
      \end{tabular}
  }
  \caption{\textbf{Comparisons in in-distribution scenarios on MegaDepth \cite{li2018megadepth}.} NG-RANSAC is trained on the same matchers in the baseline but on diffused matches in our paradigm.}
  \label{tab:ng_ransac_megadepth_performance}
\end{table}

\subsection{Ablation Study}

\subsubsection{Compatibility with learning-based RANSACs}
Recall that we conduct experiments using NG-RANSAC by default. To shed more light on the compatibility with other learning-based RANSACs, we repeat the experiments in \cref{sec:ood} and \cref{sec:id}, employing another representative learning-based RANSAC, $\nabla$-RANSAC \cite{wei2023generalized}. We report the results on ScanNet in \Cref{tab:gd_ransac_scannet} and \Cref{tab:gd_ransac_scannet_performance}. As demonstrated in these tables, our method maintains the superiority when combined with $\nabla$-RANSAC. The model trained on the diffused matches exhibits better generalization to out-of-distribution data, and its results are comparable to the ones trained and tested on the same matchers. These experiments highlight Monte Carlo diffusion as a universal enhancer for learning-based RANSACs, improving the generalization ability.

\begin{table}[htbp]
  \centering
  \resizebox{1.0\linewidth}{!}{
      \begin{tabular}{@{}lllll@{}}
        \toprule
        Training & Testing & AUC $@5^\circ$ & AUC $@10^\circ$ & AUC $@20^\circ$ \\
        \midrule
        LoFTR \cite{sun2021loftr} & SIFT \cite{lowe2004distinctive} & 2.7 & 4.4 & 10.5 \\
        MCD & SIFT \cite{lowe2004distinctive} & \textbf{7.6} {\color{rise}\footnotesize(+4.9)} & \textbf{15.8} {\color{rise}\footnotesize(+11.4)} & \textbf{26.2} {\color{rise}\footnotesize(+15.7)}\\
        \hline
        SIFT \cite{lowe2004distinctive}  & LoFTR \cite{sun2021loftr} & 22.0 & 41.4 & 58.8 \\ 
        MCD & LoFTR \cite{sun2021loftr} & \textbf{23.5} {\color{rise}\footnotesize(+1.5)} & \textbf{43.1} {\color{rise}\footnotesize(+1.7)} & \textbf{60.2} {\color{rise}\footnotesize(+1.4)}\\ 
        \bottomrule
      \end{tabular}
  }
  \caption{\textbf{Compatibility with $\nabla$-RANSAC \cite{wei2023generalized} in out-of-distribution data.} The evaluation is conducted employing $\nabla$-RANSAC, and the AUCs on ScanNet \cite{dai2017scannet} are reported.}
  \label{tab:gd_ransac_scannet}
\end{table}

\begin{table}[htbp]
  \centering
  \resizebox{1.0\linewidth}{!}{
      \begin{tabular}{@{}lllll@{}}
        \toprule
        Training & Testing & AUC $@5^\circ$ & AUC $@10^\circ$ & AUC $@20^\circ$ \\
        \midrule
        SIFT \cite{lowe2004distinctive} & SIFT \cite{lowe2004distinctive} & 7.3 & \textbf{16.6} & \textbf{28.5} \\
        MCD & SIFT \cite{lowe2004distinctive} & \textbf{7.6} & 15.8 & 26.2 \\
        \hline
        LoFTR \cite{sun2021loftr} & LoFTR \cite{sun2021loftr} & 23.1 & 42.8 & 60.0 \\ 
        MCD & LoFTR \cite{sun2021loftr} & \textbf{23.5} & \textbf{43.1} & \textbf{60.2} \\ 
        \bottomrule
      \end{tabular}
  }
  \caption{\textbf{Compatibility with $\nabla$-RANSAC \cite{wei2023generalized} in in-distribution scenarios.} Results on ScanNet \cite{dai2017scannet} are listed.}
  \label{tab:gd_ransac_scannet_performance}
\end{table}

\subsubsection{Monte Carlo approximation}
MCD approximates all possible distributions $\mathcal{D}_{\text{all}}$ of the raw data through Monte Carlo sampling \cite{metropolis1949monte}. A straightforward alternative is to approximate $\mathcal{D}_{\text{all}}$ by combining multiple matchers for data generation. Therefore, we construct a training dataset using both SIFT \cite{lowe2004distinctive} and LoFTR \cite{sun2021loftr} to generate correspondences between images, doubling the size of the original dataset that uses a single matcher. We employ SuperGlue (SG) \cite{sarlin2020superglue} to establish correspondences during testing, treating it as out-of-distribution data. The experimental results on ScanNet are shown in \Cref{tab:monte_carlo_approximation}. Our MCD consistently outperforms the baseline, demonstrating better generalization to out-of-distribution data. While integrating multiple matchers expands the training dataset and covers more distribution types, fixed matchers cannot sufficiently explore $\mathcal{D}_{\text{all}}$. In contrast, our method is independent of specific mathers, and the inherent randomness of the Monte Carlo diffusion module enables diverse sampling from $\mathcal{D}_{\text{all}}$. This advantage thereby leads to improved generalization.

\begin{table}[htbp]
  \centering
  \resizebox{1.0\linewidth}{!}{
      \begin{tabular}{@{}lclll@{}}
        \toprule
        Training & Testing & AUC $@5^\circ$ & AUC $@10^\circ$ & AUC $@20^\circ$ \\
        \midrule       
        SIFT \cite{lowe2004distinctive}+LoFTR \cite{sun2021loftr} & SG \cite{sarlin2020superglue} & 15.5 & 33.4 & 51.8 \\
        MCD & SG \cite{sarlin2020superglue} & \textbf{17.4} \,{\color{rise}{\footnotesize(+1.9)}} & \textbf{35.5} \,{\textcolor{rise}{\footnotesize(+2.1)}} & \textbf{54.0} \,{\textcolor{rise}{\footnotesize(+2.2)}} \\
        \bottomrule
      \end{tabular}
  }
 \caption{\textbf{Effectiveness of Monte Carlo approximation.} NG-RANSAC is trained on SIFT and LoFTR as a baseline. The baseline and the model trained on our diffused matches are tested on SuperGlue (SG) \cite{sarlin2020superglue} that indicates out-of-distribution data. AUCs on ScanNet \cite{dai2017scannet} are reported.}
 \label{tab:monte_carlo_approximation}
\end{table}


\subsubsection{Multi-Stage Randomization}
The randomness in the MSR module is controlled through three parameters, i.e., timestep $t$, diffusion ratio $r$, and noise scale $s$. To assess the contribution of each component, we progressively incorporate them into our framework. We train NG-RANSAC on diffused matches generated by each variant, and \Cref{tab:multi_layer} lists the ablation results on ScanNet. The ablation starts with randomized $t$ while keeping $r=0.5$ and $s=0.1$ fixed. The inclusion of randomized $r$ and $s$ improves performance on both SIFT and LoFTR during testing, increasing AUC $@20^\circ$ by $19.6\%$ and $4.0\%$ on SIFT and LoFTR, respectively. These parameters introduce different types of randomness: $t$ determines the distribution of the noised data, $r$ controls the outlier ratio in diffused matches, and $s$ defines the noise level. Their combined effects enhance the diversity of generated data, leading to an optimal solution in our pipeline, as demonstrated by the experimental results.

\begin{table}[htbp]
  \centering
  \resizebox{1.0\linewidth}{!}{
    \begin{tabular}{@{}cccllll@{}}
      \toprule
      $t$ & $r$ & $s$ & Testing & AUC $@5^\circ$ & AUC $@10^\circ$ & AUC $@20^\circ$°\\
      \midrule
      \checkmark & - & - & SIFT \cite{lowe2004distinctive} & 1.1 & 3.2 & 6.6 \\
      \cmidrule(lr){4-7}
      \checkmark & - & - & LoFTR \cite{sun2021loftr} & 20.1 & 39.0 & 56.8 \\
      \midrule
      \checkmark & \checkmark & - & SIFT \cite{lowe2004distinctive} & 
      4.9 \,{\color{SIFT_rise}\footnotesize(+3.8)} & 
      10.6 \,{\color{SIFT_rise}\footnotesize(+7.4)} & 
      18.3 \,{\color{SIFT_rise}\footnotesize(+11.7)} \\
      \cmidrule(lr){4-7}
      \checkmark & \checkmark & - & LoFTR \cite{sun2021loftr} & 
      21.7 \,{\color{loftr_rise}\footnotesize(+1.6)} & 
      42.2 \,{\color{loftr_rise}\footnotesize(+3.2)} & 
      59.8 \,{\color{loftr_rise}\footnotesize(+3.0)} \\
      \midrule
      \checkmark & \checkmark & \checkmark & SIFT \cite{lowe2004distinctive} & 
      7.6 \,{\color{SIFT_rise}\footnotesize(+2.7)} & 
      16.2 \,{\color{SIFT_rise}\footnotesize(+5.6)} & 
      26.2 \,{\color{SIFT_rise}\footnotesize(+7.9)} \\
      \cmidrule(lr){4-7}
      \checkmark & \checkmark & \checkmark & LoFTR \cite{sun2021loftr} & 
      22.4 \,{\color{loftr_rise}\footnotesize(+0.7)} & 
      42.7 \,{\color{loftr_rise}\footnotesize(+0.5)} & 
      60.8 \,{\color{loftr_rise}\footnotesize(+1.0)} \\
      \bottomrule
    \end{tabular}
  }
  \caption{\textbf{Ablation study on MSR.} $t$, $r$, and $s$ represent timestep, diffusion ratio, and noise scale, respectively. Each of these components introduces different types of randomness in MSR. We progressively add them to our pipeline, and the results on ScanNet \cite{dai2017scannet} are shown.}
  \label{tab:multi_layer}
\end{table}

\subsubsection{Comparison With Traditional RANSAC}
In addition to learning-based RANSACs, we conduct comparisons between our method and traditional RANSAC variants. Specifically, we maintain the experimental setup, training NG-RANSAC on diffused matches as our method. We evaluate the trained model and the handcraft methods, RANSAC and MAGSAC++, on SIFT and LoFTR. The results on ScanNet are presented in \Cref{tab:Comparison With RANSACS On ScanNet}. Traditional methods exhibit limited performance on SIFT, resulting in an AUC $@5^\circ$ of 0.6 for RANSAC and 0.8 for MAGSAC++. As SIFT typically produces noisy initial matches with numerous outliers, this finding indicates that handcraft RANSACs are sensitive to outliers, limiting their effectiveness in applications requiring robust parametric model estimation from low-quality data. In contrast, the learning-based RANSAC trained based on our Monte Carlo match diffusion module demonstrates superior adaptability across scenarios with varying data quality.

\begin{table}[htbp]
  \centering
  \resizebox{1.0\linewidth}{!}{
      \begin{tabular}{@{}lllll@{}}
        \toprule
        Method & Testing & AUC $@5^\circ$ & AUC $@10^\circ$ & AUC $@20^\circ$ \\
        \midrule
        RANSAC~\cite{fischler1981random} & SIFT \cite{lowe2004distinctive} & 0.6 & 2.1 & 10.5 \\
        MAGSAC++~\cite{barath2020magsac++} & SIFT \cite{lowe2004distinctive} & 0.8 & 2.3 & 5.4 \\
        Ours & SIFT \cite{lowe2004distinctive} & \textbf{7.6} {\textcolor{rise}{\footnotesize(+6.8)}} & \textbf{16.2} {\textcolor{rise}{\footnotesize(+13.9)}}& \textbf{26.2} {\textcolor{rise}{\footnotesize(+20.8)}} \\
        \hline
        RANSAC~\cite{fischler1981random} & LoFTR \cite{sun2021loftr} & 20.4 & 39.4 & 58.1 \\ 
        MAGSAC++~\cite{barath2020magsac++} & LoFTR \cite{sun2021loftr} & 21.6 & 41.1 & 59.3 \\ 
        Ours & LoFTR \cite{sun2021loftr} & \textbf{22.4} {\textcolor{rise}{\footnotesize(+0.8)}}& \textbf{42.7} {\textcolor{rise}{\footnotesize(+1.6)}} & \textbf{60.8} {\textcolor{rise}{\footnotesize(+1.5)}}\\ 
        \bottomrule
      \end{tabular}
  }
  \caption{\textbf{Comparison with handcrafted RANSAC variants on ScanNet \cite{dai2017scannet}.} We train NG-RANSAC based on MCD and compare the model with RANSAC and MAGSAC++.}
  \label{tab:Comparison With RANSACS On ScanNet}
\end{table}

\section{Conclusion}
\label{sec:conclusion}
In this paper, we have presented a Monte Carlo diffusion mechanism that enhances the generalization ability of learning-based RANSAC. Unlike existing methods that suffer from overfitting to specific data distributions, our approach decouples training from fixed data sources by simulating diverse noise patterns through a diffusion-driven process. By incorporating Monte Carlo sampling, we inject randomness at multiple stages of the diffusion module, further increasing data diversity and robustness. We have evaluated our method on ScanNet and MegaDepth, demonstrating that learning-based RANSAC trained based on Monte Carlo diffusion achieves significantly better generalization when tested on out-of-distribution data. Our method also maintains competitive performance in in-distribution scenarios, ensuring broad applicability. Additionally, extensive ablation studies have confirmed the compatibility with different learning-based RANSAC variants and the effectiveness of key components in our framework.
\clearpage
{
    \small
    \bibliographystyle{ieeenat_fullname}
    \bibliography{egbib}
}


\end{document}